\documentclass[lettersize,journal]{IEEEtran}
\usepackage{amsmath,amsfonts}
\usepackage{algorithmic}
\usepackage{algorithm}
\usepackage{array}
\usepackage[caption=false,font=normalsize,labelfont=sf,textfont=sf]{subfig}
\usepackage{textcomp}
\usepackage{stfloats}
\usepackage{url}
\usepackage{verbatim}
\usepackage{graphicx}
\usepackage{cite}
\usepackage{multirow}
\usepackage{pifont}

\begin{document}

\title{Bi-ACT: Bilateral Control-Based Imitation Learning via Action Chunking with Transformer}


\author{
\IEEEauthorblockN{Thanpimon Buamanee$^\dag$$^{1}$, Masato Kobayashi$^\dag$$^{2*}$,  Yuki Uranishi$^{2}$, Haruo Takemura$^{2}$}
\thanks{
$^{\dag}$Equal Contribution,\\
$1$ School of Engineering Science, Osaka University, Toyonaka, Osaka 560-0043, Japan,
$2$ Cybermedia Center, Osaka University, Toyonaka, Osaka, 560-0043 Japan,
$^{*}$Email:kobayashi.masato.cmc@osaka-u.ac.jp
}
}



\maketitle

\begin{abstract}
Autonomous manipulation in robot arms is a complex and evolving field of study in robotics.
This paper proposes work stands at the intersection of two innovative approaches in the field of robotics and machine learning. Inspired by the Action Chunking with Transformer (ACT) model, which employs joint location and image data to predict future movements, our work integrates principles of Bilateral Control-Based Imitation Learning to enhance robotic control. Our objective is to synergize these techniques, thereby creating a more robust and efficient control mechanism.
In our approach, the data collected from the environment are images from the gripper and overhead cameras, along with the joint angles, angular velocities, and forces of the follower robot using bilateral control. The model is designed to predict the subsequent steps for the joint angles, angular velocities, and forces of the leader robot. This predictive capability is crucial for implementing effective bilateral control in the follower robot, allowing for more nuanced and responsive maneuvering.
\end{abstract}


\section{Introduction}
\IEEEPARstart{I}{n} recent years, the paradigm of robot control has shifted significantly, with an increasing emphasis on learning from human demonstration, often referred to as behavior cloning or imitation learning\cite{IMIB2022yang, IMI2023xu, IMI2022ding, IMI2018yang}.
This approach, rooted in observing and replicating human actions, offers a more intuitive and adaptable framework for robotic control, particularly in complex and unstructured environments.
In imitation learning, there are generally three steps: data collection by experts, learning from the collected data, and autonomous operation using the learned model. Especially in imitation learning, high-quality expert data, the architecture of the learning model, and a robot system design suitable for imitation learning are required\cite{IMI2023zhou, IMI2023franzese}.

In the field of IL, the method of collecting robot data is a crucial factor\cite{lfd2009argall, lfd2022mukherjee}.
Data acquisition frequently involves the utilization of teleoperated systems.
These systems encompass a range of devices including, but not limited to, virtual reality headsets coupled with hand-tracking mechanisms\cite{tre2018Zhang}, smartphones\cite{tre2021tung}, keyboard inputs\cite{tre2018fan}, and leader-follower system\cite{act2023zhao, mact2024fu, ge2023swu}.
Especially, the ALOHA\cite{act2023zhao}, Mobile ALOHA\cite{mact2024fu}, and GELLO\cite{ge2023swu} systems have shown particularly notable results in data collection methods for imitation learning. These are leader-follower type systems that collect robot joint angles and image data. The ALOHA system is based on position control for data collection, so it does not handle force information and therefore cannot discern the hardness of objects.
Therefore, data collection using bilateral control, which can handle both position and force information, has also garnered attention\cite{IMIB2022sakaino}.

\begin{figure}[t]
  \begin{center}
    \scalebox{0.35}{
        \includegraphics{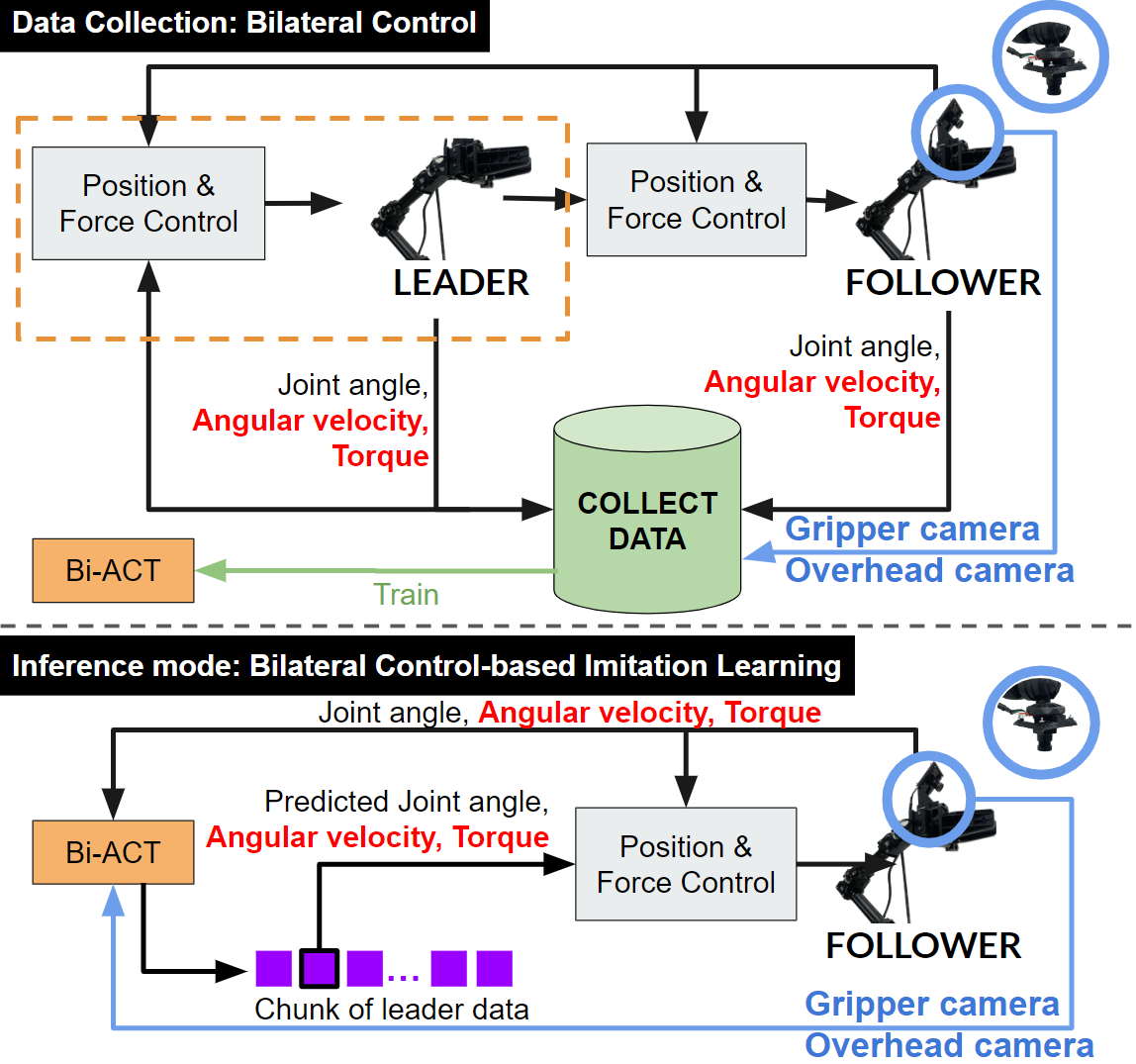}}    
  \caption{Overview of Bilateral Control-Based Imitation Learning via Action Chunking with Transformer (Bi-ACT)}
  \label{fig:overview}
\end{center}
\end{figure}

Imitation learning enables efficient skill acquisition by learning models from demonstration data collected by experts. However, even with high-quality data, acquiring skills can be challenging if the model architecture is suboptimal. Conventionally, RNN and LSTM\cite{lstm} have been employed for handling time-series data, and imitation learning methods utilizing Transformers have also been reported.
Small errors in the predicted action can incur large differences in the state, exacerbating the “compounding error” problem of imitation learning\cite{con2021ke}.
However, Action Chunking with Transformer (ACT)\cite{act2023zhao}, trained with data collected via ALOHA, has overcome these challenges.
In ACT, the policy predicts the target joint positions for the next $k$ time steps, not just one step at a time. This reduces the task's effective horizon by a factor of $k$, thereby diminishing the accumulation of errors.
Predicting sequences of actions is also beneficial in addressing temporally correlated confounders, such as pauses during demonstrations, which are difficult to model with Markovian single-step policies \cite{con2022swamy}.

Consequently, we were inspired by the imitation learning model employing the ACT method and a remote control technique capable of collecting positional and force information, known as bilateral control.
The ALOHA had been collecting data without utilizing force control and had not incorporated force information into its learning process\cite{act2023zhao, mact2024fu}.
Moreover, the imitation learning method based on bilateral control, proposed by Sakaino\cite{IMIB2022sakaino}, utilizes both positional and force information; however, most of these models predominantly employ Long Short-Term Memory (LSTM)\cite{lstm}.
Thus this paper proposes the "Bilateral Control-Based Imitation Learning via Action Chunking with Transformer (Bi-ACT)".
Our approach, Bi-ACT, builds upon these foundations, aiming to create a more robust and responsive control system for autonomous robotic arms. The key innovation lies in the integration of bilateral control principles with imitation learning strategies. Bilateral control, characterized by its two-way communication and feedback loop, allows for a more nuanced and synchronized interaction between the leader (human operator) and the follower (robot). This interaction is crucial for tasks requiring precision and adaptability, such as delicate manipulation or operation in dynamic environments. Fig.~\ref{fig:overview} provides a schematic representation of our method.

The main contributions of this paper are twofold:
\begin{itemize}
\item This paper proposes a novel approach to "Bilateral Control-Based Imitation Learning via Action Chunking with Transformer", hereafter referred to as Bi-ACT.
Bi-ACT is trained using joint angles, joint velocities, torque, and images. The robot operates via bilateral control based on position and force information. This enables adaptation to the hardness and weight of objects, which was not possible with only position control.
\item Bi-ACT takes joint angles, joint velocities, torque, and images as inputs, and produces joint angles, joint velocities, and torque as outputs. As a multimodal data handling robotic system, the output from the model utilizes Action Chunks, achieving a frequency of 100Hz. This enables fast and robust motion generation in an imitation learning system that handles image data. 
\end{itemize}
Finally, the effectiveness of Bi-ACT has been validated through extensive real-world experiments involving pick-and-place and put-in-drawer tasks with objects of varying hardness.

This paper consists of eight sections including this one.
Sections II and III explain the Related works, Control System.
Section IV proposes BI-ACT.
In Sections V, experimental results are shown to confirm the usefulness of the proposed method.
Section VI concludes this paper.

\section{Related Works}
\subsection{Bilateral Control-Based Imitation Learning}
\begin{figure}[t]
  \begin{center}
    \scalebox{0.5}{
        \includegraphics{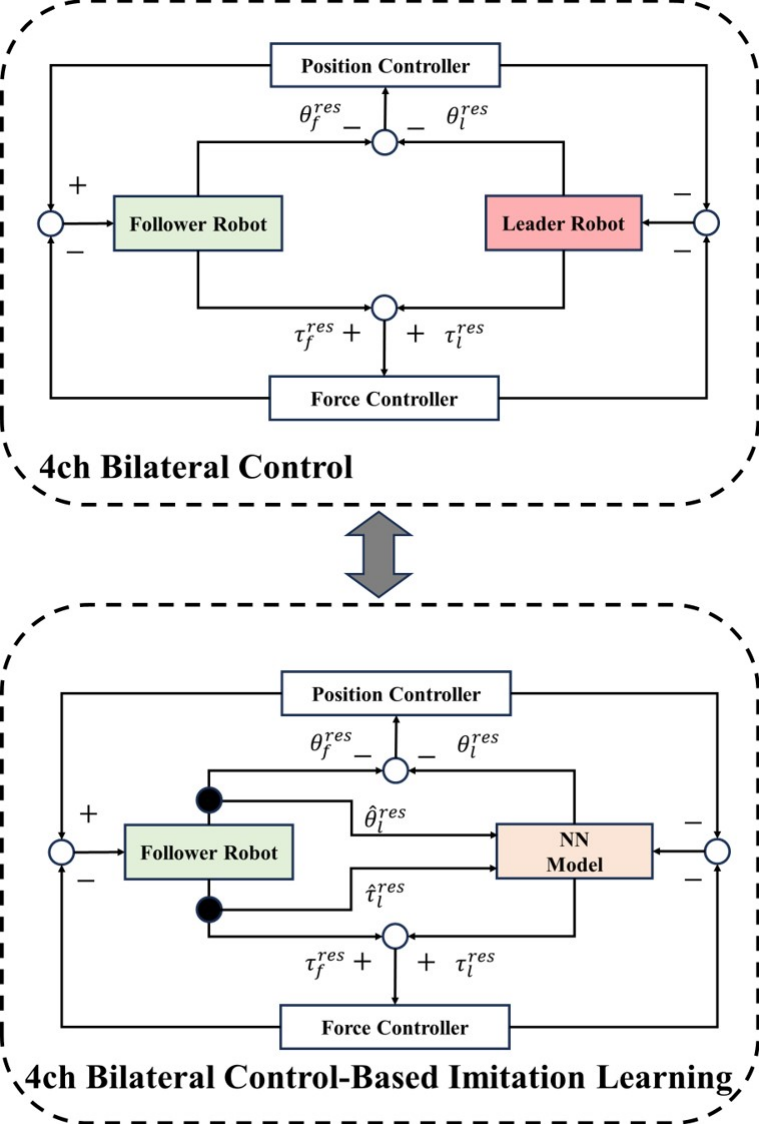}}    
  \caption{Block Diagram of Four-channel Bilateral Control and Four-channel Bilateral Control-Based Imitation Learning}
  \label{fig:4ch}
\end{center}
\end{figure}
Imitation learning based on bilateral control, capable of collecting position and force information, has been proposed by Sakaino and others\cite{IMIB2018adachi}~-\cite{IMIB2023kobayashi}.
In imitation learning through bilateral control, this control method is utilized for data collection. Bilateral control involves the remote operation of a follower robot in the environment, guided by a leader robot controlled by a human.
This is achieved through position tracking and the use of action-reaction principles. Various tasks have been accomplished using these bilateral control-based imitation learning methods.
For instance, Adachi et. al. reported on using a robot to draw a line along a ruler \cite{IMIB2018adachi}.
Sakaino et. al. reported on slicing cucumber task \cite{IMIB2022sakaino}.
Yamane et. al. reported on nail hand for bilateral control based imitation learning \cite{IMIB2023yamane}.
These methods have previously implemented imitation learning through bilateral control using LSTM\cite{lstm}.
Additionally, Kobayashi et. al. have proposed imitation learning with bilateral control utilizing Transformers\cite{IMIB2023kobayashi,TRANS2017vaswani}.
Such methods, which facilitate the collection of learning data and autonomous robot operation at a speed comparable to humans, are extremely valuable. However, most of these methods do not use image data for robot operation, which limits their robustness to changes in the operating environment. Addressing this gap, our study introduces Bi-ACT, a method combining ACT\cite{act2023zhao} with bilateral control-based imitation learning that incorporates image data.

\subsection{Action Chunking with Transformer}
The inception of Action Chunking with Transformers (ACT) represented a significant leap in behavior cloning algorithms in robotics\cite{act2023zhao, mact2024fu}.
ACT utilizes a Conditional Variational Autoencoder (CVAE) to model diverse scenes, in conjunction with a transformer capable of predicting sequences of actions (or 'chunks') from multimodal inputs.
This method casts the predicted actions as goal states for the manipulator, allowing for a temporal aggregation of actions across multiple time steps through a weighted average. Such aggregation helps mitigate the issues of compounding errors and unpredictable responses in out-of-distribution states. While there is still a possibility of selecting erroneous actions, the inclusion of correctly predicted actions from earlier time steps serves to moderate the final action chosen.
Building on the foundation laid by ACT, the One ACT Play methodology has emerged as a notable enhancement in this domain\cite{oneact2023george}.
One ACT Play differentiates itself by using the robot's end-effector position and posture, along with images, as inputs, in contrast to ACT's reliance on joint angles and images. This shift enables a more intuitive and direct interaction with the robot's environment. Additionally, One ACT Play introduces an innovative strategy of augmenting single demonstrations, thereby simplifying and streamlining the data collection process.
Despite these advancements, both ACT and One ACT Play methodologies do not incorporate force information during data collection or robot operation.
This limitation is noteworthy as the integration of both positional and force information through bilateral control could further enhance the effectiveness of imitation learning. Incorporating bilateral control, which involves simultaneous management of both position and force, has the potential to significantly improve the quality of data collection and increase the success rate of autonomous tasks.
Thus, Bi-ACT promises a more comprehensive understanding and manipulation of the robot's environment, leading to more precise and adaptable task execution.

\section{Control System}
\subsection{Controller}
\begin{figure}[t]
  \begin{center}
    \scalebox{0.25}{
        \includegraphics{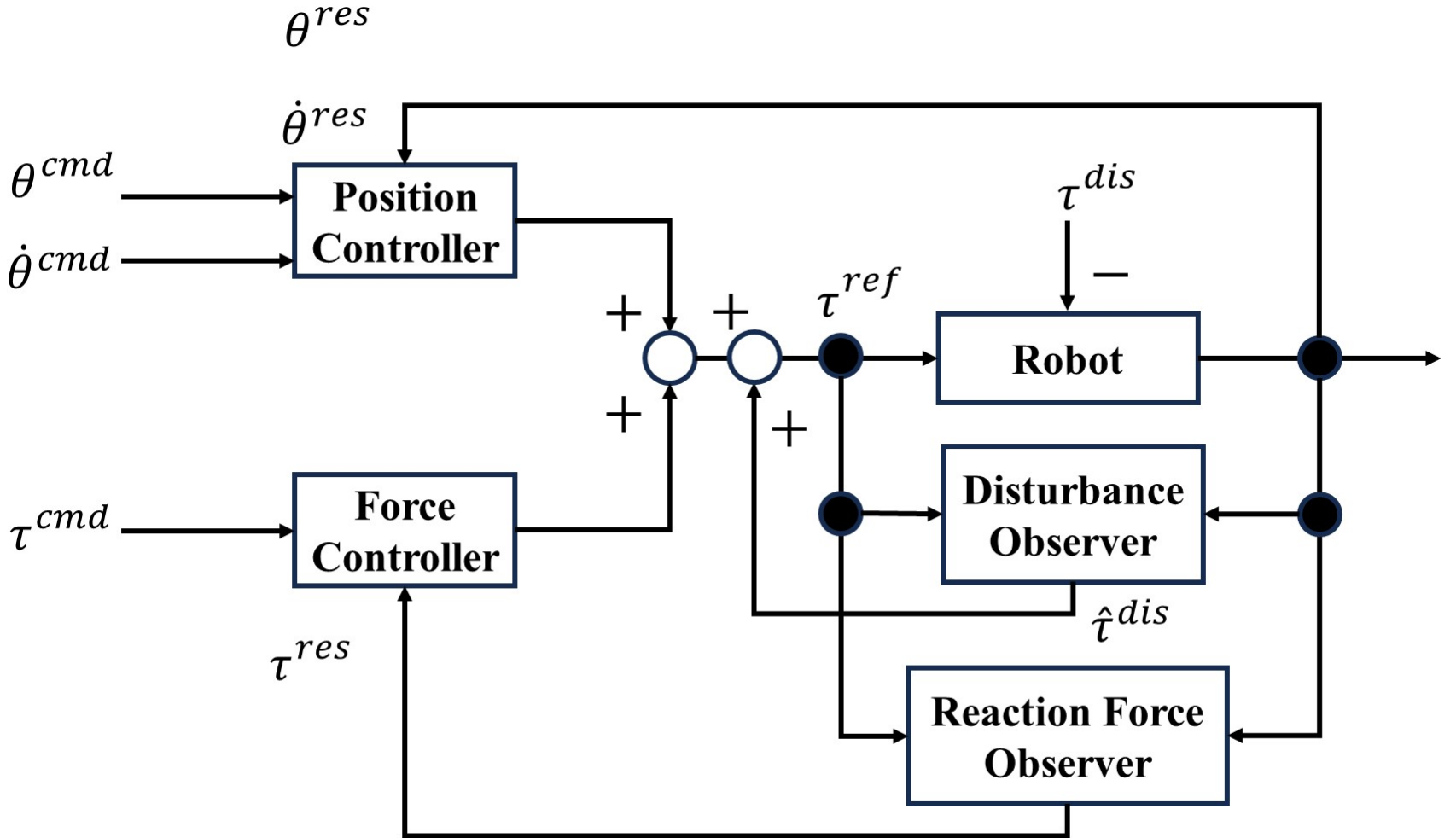}}    
  \caption{Block Diagram of Control System}
  \label{fig:robot_system}
\end{center}
\end{figure}

The controller design adopted a control of position and force for each axis, as shown in Fig.~\ref{fig:robot_system}.
Angle information was obtained from encoders, and angular velocity was calculated by differentiating this information. The disturbance torque $\hat{\tau}^{dis}$ was calculated using a disturbance observer (DOB) \cite{DOB}, and the torque response value $\tau^{res}$ was estimated using a force reaction observer (RFOB) \cite{RFOB}.
\subsection{Bilateral Control}
The fundamental principle of bilateral control is sharing position, force, or other information between the operator and the control target.
The control goals of the bilateral control are summarized as follows:
\begin{equation}
\theta_l - \theta_f = 0
\label{eq:position} 
\end{equation}
\begin{equation}
\tau_l + \tau_f = 0
\label{eq:force}
\end{equation}
where $\theta$ and $\tau$ represents the joint angle and torque.
The subscript $\bigcirc_l$ represents the leader system, and $\bigcirc_f$ represents the follower system.
This allows the operator to perform intuitive control over the control target.
Specifically, bilateral control is achieved by satisfying (\ref{eq:position}), representing position tracking between systems, and (\ref{eq:force}), representing the action-reaction relationship of forces.

The control block diagram for a follower in autonomous operation is depicted on the right side of Fig.~\ref{fig:4ch}. During autonomous operation, a trained neural network (NN) model substitutes for the leader. The follower robot is controlled by receiving predictions of the leader's response, as forecasted by the model, and using these predictions as commands. This trained NN model forecasts the leader's next response by considering both the follower's received response and the model's previous step prediction.

\section{Bi-ACT: Bilateral Control-Based Imitation Learning via Action Chunking with Transformer}
\subsection{Overview}
Our proposed work stands at the intersection of two innovative approaches in the field of robotics and machine learning. Drawing inspiration from the groundbreaking methodology presented in the ACT research, which utilizes two types of input data: joint locations and image data, to predict subsequent movements, we also incorporate principles of Bilateral Control-Based Imitation Learning. Our aim is to synergies these methodologies to forge a more robust and efficient approach for robotic control.\\

In our approach, the data collected from the environment are images from the gripper and overhead cameras, along with the joint angles, angular velocities, and forces of the follower robot using bilateral control. The model is designed to predict the subsequent steps for the joint angles, angular velocities, and forces of the leader robot. This predictive capability is crucial for implementing effective bilateral control in the follower robot, allowing for more nuanced and responsive maneuvering.

\subsection{Data Collection}
Data collection was performed using Bilateral Control, which allowed the user to feel the follower robot's environment while controlling the leader robot thus increasing the demonstration data quality.  Both leader and follower robot arm's joint angles, angular velocities, and forces, along with images collected from overhead and gripper cameras, were recorded while completing the desired task. With the increase on the force as an input, the difference in object's weight and texture will be take in consideration when training the model.
\subsection{Learning Architecture}
\begin{figure*}[t]
  \begin{center}
    \scalebox{0.33}{
        \includegraphics{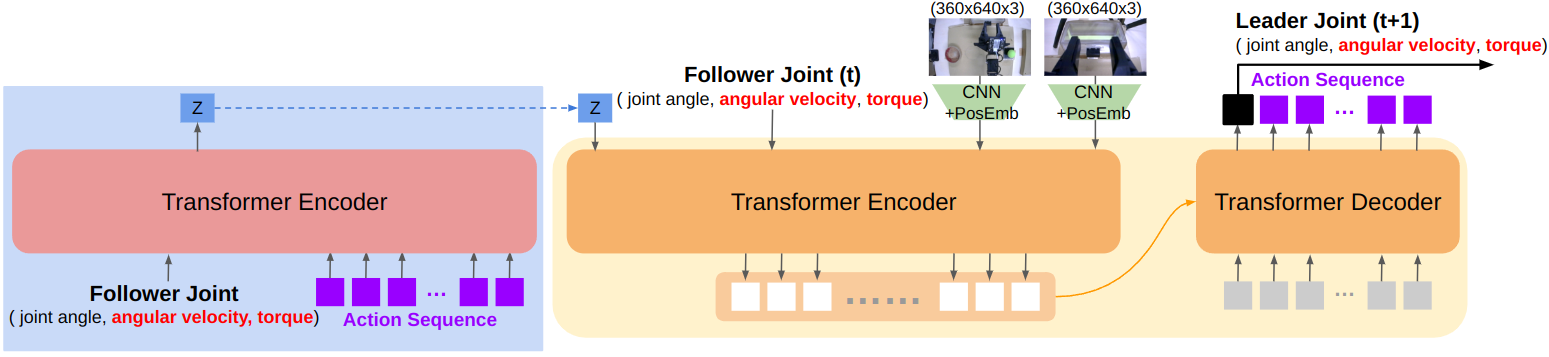}}    
  \caption{Model Architecture: Bilateral Control-Based Imitation Learning via Action Chunking with Transformer}
  \label{fig:Proposed-Model}
\end{center}
\end{figure*}

With the purpose of improving the comprehensibility of the environment in the study, our method were based on Action Chunking with Transformer, with an addition of a novel dimension to the input and output data. Along with joint angle and images data which are previously used in the original work, we increased angle velocity and force to the input data. A schematic representation of our network structure is provided in Fig.~\ref{fig:Proposed-Model}.

From the architecture, it is evident that the model receives inputs as two RGB images, each captured at a resolution of 360 x 640, one from the follower's gripper and the other from an overhead perspective. In addition, the model processes the current follower's joint data, which consists of three types of data (angle, angular velocity, and force) across five joints, forming a 15-dimensional vector in total. Utilizing action chunking, the policy generates an $k$ x 15 tensor, representing the leader's next actions over $k$ time steps. The leader's actions for these time steps are then conveyed to the controller, which determines the required current for the joints in the follower robot, enabling it to move in the specified direction.

\subsection{Execution to Robot Arm}
In this paper, as the action data output by the model in each time step consists of 3 data: joint angle, velocity and torque of each joint, it will be transformed into the requisite current for each joint to achieve the desired state, facilitated by calculations performed by the bilateral control system. 
Given the requisite sensitivity and precision in current control, it is essential for the robot to operate at a high frequency to ensure optimal performance. Consequently, we have set the action result update frequency at 100 Hz.
The model is executed every $k$ time steps, during which it generates predictions for the subsequent $k$ time steps.
These predictions enable the robot to function effectively at the target 100 Hz frequency.
As a result, the model can autonomously perform tasks such as pick-and-place with both familiar and unfamiliar objects, demonstrating its proficiency in adapting to new environments. Furthermore, the robot can execute prolonged tasks, such as placing items in a drawer, which showcases its efficiency in recognizing and accurately performing different segments of a task.
\section{Experiments}
\subsection{Hardware}
\begin{figure}[t]
  \begin{center}
    \scalebox{0.25}{
        \includegraphics{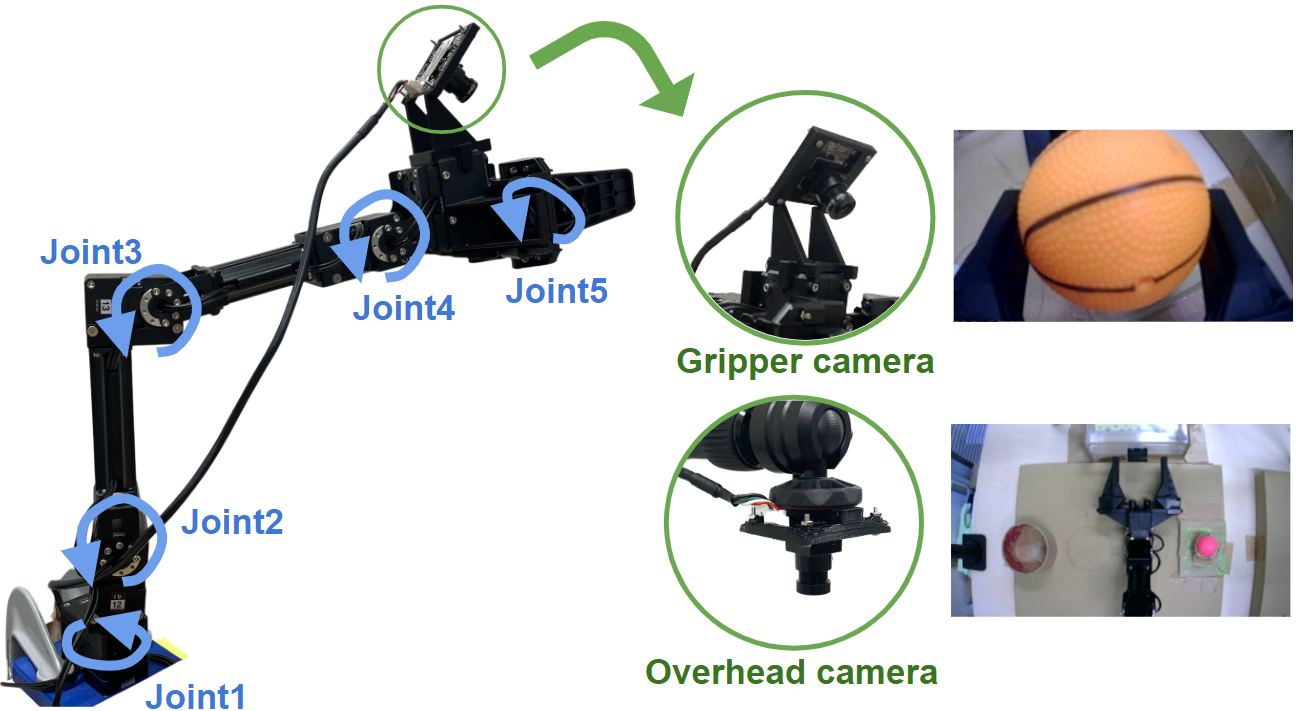}}    
  \caption{Definition of Robot and Camera View}
  \label{fig:Hardware}
\end{center}
\end{figure}

As shown in Fig.~\ref{fig:Hardware}, the OpenMANIPULATOR-X robotic arm, manufactured by ROBOTIS, was utilized for experiments. This robot possesses 4 Degrees of Freedom (DOF), enabling movement in multiple directions, and an additional DOF for the gripper. The control cycle period was set to 1000Hz for precise movement. Moreover, RGB Camera were placed in the overhead area and on the gripper area of the follower robot to record observations.

\subsection{Environment Setting}
\begin{figure}[t]
  \begin{center}
    \scalebox{0.2}{
        \includegraphics{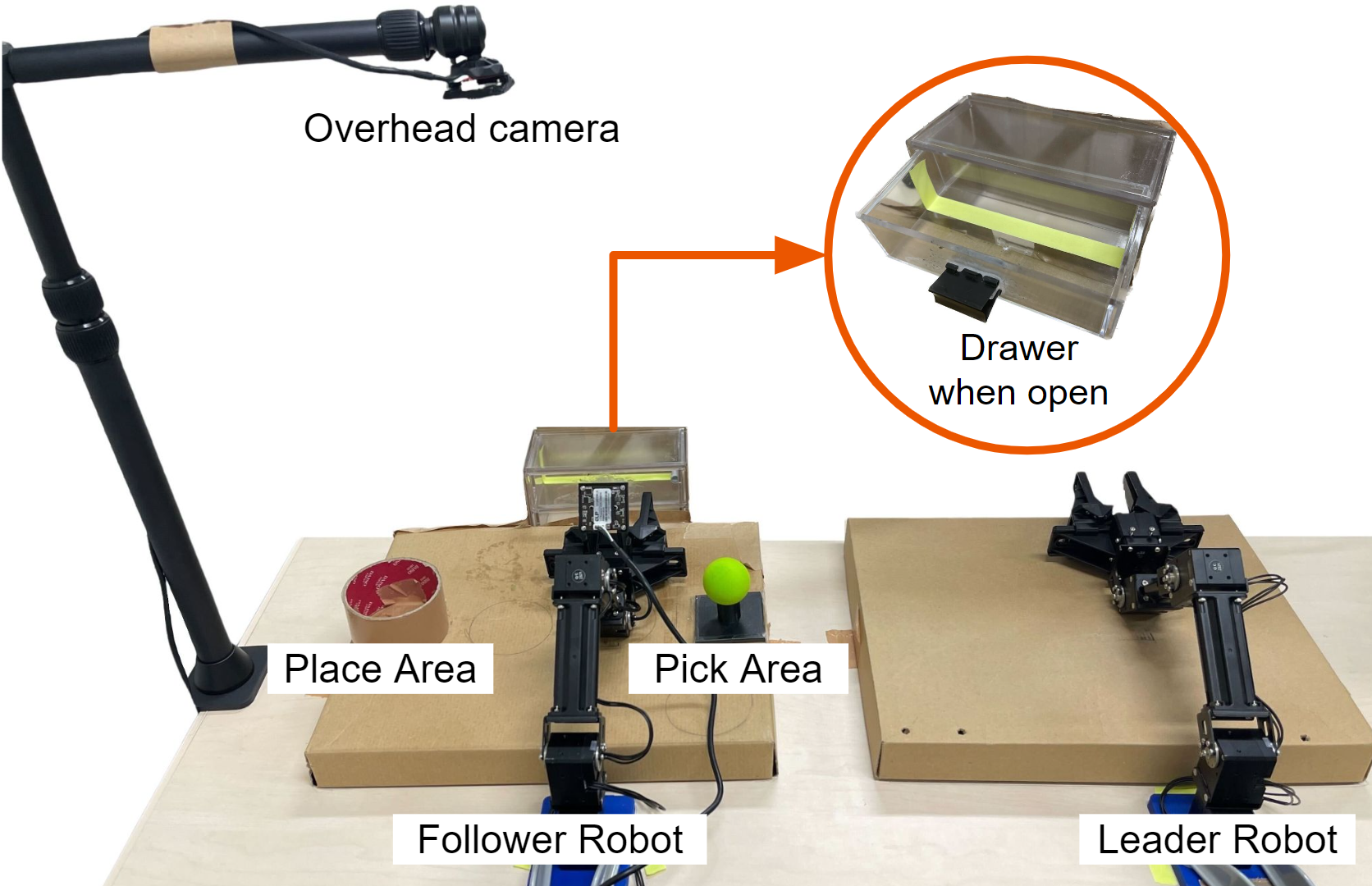}}    
  \caption{Experimental Environments Setup}
  \label{fig:Environment-Setup}
\end{center}
\end{figure}

There were two experimental cases: Case 1 was 'Pick-and-Place,' and Case 2 was 'Put-in-Drawer.'
In the environment setup, two robotic units designated as the leader and the follower, were positioned adjacent to each other as delineated in Fig.~\ref{fig:Environment-Setup}. The experimental environment was arranged on the side of the follower robot, which was the designated site for task execution.

\begin{figure}[t]
  \begin{minipage}{0.49\hsize}
    \begin{center}
      \scalebox{0.19}{
        \includegraphics{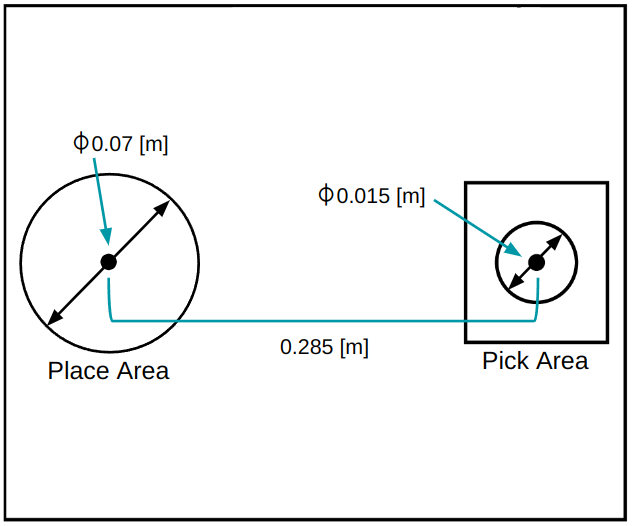}}
      \vspace{2mm}
      {\begin{center} (a) Pick-and-Place\end{center}}
    \end{center}
  \end{minipage}
  \begin{minipage}{0.49\hsize}
    \begin{center}
      \scalebox{0.16}{
        \includegraphics{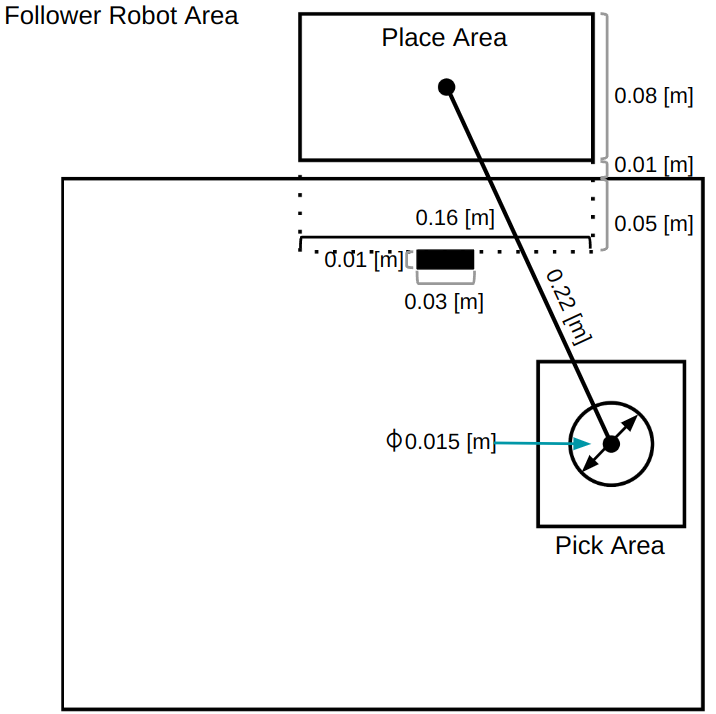}}
      {\begin{center} (b) Put-in-drawer\end{center}}
    \end{center}
  \end{minipage}
  \caption{Image of Experimental Environments}
  \label{fig:Setting}
\end{figure}%
As illustrated in Fig.~\ref{fig:Setting}, we evaluated Bi-ACT's performance on two manipulation tasks: Pick-and-Place and Put-in-Drawer. Both tasks were performed to test the performance of the robot's control system, with the first task focusing on how the robot performs with untrained data, and the second task designed to evaluate the proposed model's performance on long-duration tasks.

\begin{figure}[t]
  \begin{center}
    \scalebox{0.13}{
        \includegraphics{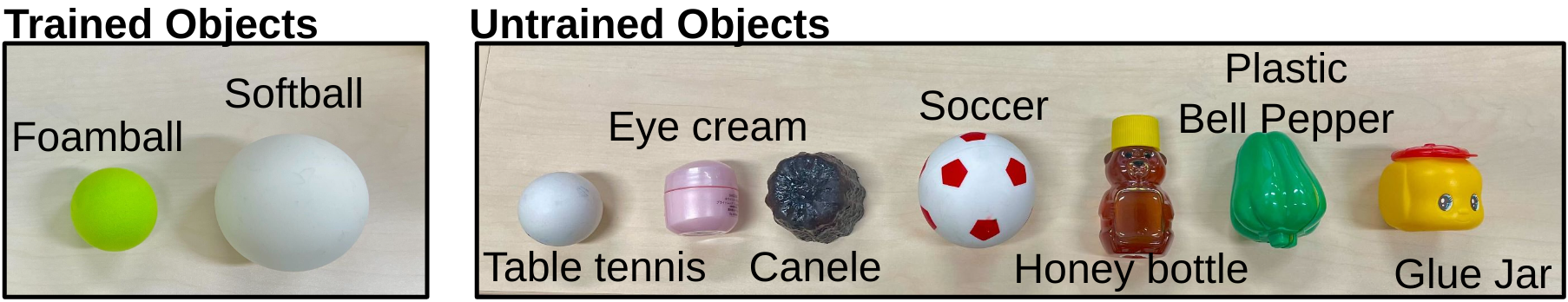}}    
  \caption{Image of Trained and Untrained Objects}
  \label{fig:Target-Object}
\end{center}
\end{figure}

For the initial task of 'Pick-and-Place' as shown in Fig.~\ref{fig:Setting}(a), the objective was for the gripper to accurately pick up objects of various shapes, weights, and textures from the pick area and then place them within the place area. The pick area was configured as a square equipped with a stabilizing stand with dimensions of 0.015 [m], serving as the initial placement point for the object. In contrast, the place area was a circular area with a diameter of 0.07 [m], established as the endpoint for object deposition. The spatial separation between these two areas was precisely measured at 0.285 [m].
We used a foam ball and a softball during the data collection phase. In testing the model, these two objects, along with seven untrained objects - a table tennis ball, an eye-cream package, Canele, a soccer ball, a plastic bell pepper, a honey bottle and a glue jar - were used, as shown in Fig.~\ref{fig:Target-Object}. The gripper's task was to transport these objects without dropping them during transit. Dropping an object outside of the designated place area was considered a task failure.

Fig.~\ref{fig:Setting}(b) displays the layout of the second task, 'Put-in-Drawer', which involved moving an object from the pick area to the drawer. The pick area, configured similarly to the first task as a square with a stabilizing stand of 0.015 [m] dimensions, served as the initial placement point for the object. The Drawer area designated for this task was 0.16m wide and featured a 0.01x0.03 [m] handle, which the robot had to grasp and pull to open the drawer. A yellow marker was placed inside the drawer, precisely 0.035 [m] from its end, to prevent the drawer from derailing and to provide a visual boundary for the robotic arm and human operators.
In the 'Put-in-Drawer' task of data collection, the robot's first challenge was to securely grasp the small and friction-prone handle of the drawer. Due to the handle's size and associated friction, even a minor error in applying force during the opening and closing process could lead to the drawer failing to open, resulting in a task failure. Additionally, opening the drawer too little would prevent the ball from being placed inside, while opening it too much could cause derailing, making the drawer unable to close properly.

Subsequently, the robot's task mirrored the first task 'Pick-and-Place': it navigated to the pick area, grasped an object, and then deposited it into the designated area inside the drawer. The final step involved the robot closing the drawer by exerting pressure on the bottom of the handle. This action enabled the drawer's derail area to ascend, allowing for a smooth closure.

\begin{table}[t]
  \caption{Detail of Trained and Untrained Objects}
    \scalebox{0.9}{
\begin{tabular}{ccccc}

\hline
 &   Training Data & Size [mm] &  Hardness &  Weight [g]        \\ \hline \hline
Foam ball  & \ding{52} & 40 & High & 3 \\ 
softball  & \ding{52} & 66 &  Very Low & 30 \\ 
Table tennis  & \ding{54} & 40 &  Very High & 2 \\ 
Eye cream   & \ding{54} & 40 &  Very High & 24 \\ 
Canele   & \ding{54} & 50 &  Low & 86 \\ 
Soccer   & \ding{54}& 61 &  Low & 21 \\ 
Honey bottle  & \ding{54} & 40 &  Very High & 79 \\ 
Plastic Bell Pepper  & \ding{54}& 49 &  Very High & 15 \\ 
Glue Jar   & \ding{54}& 45 &  Medium & 63 \\ \hline
\end{tabular}
}
\label{tbl:data_obj}
\end{table}

\begin{figure*}[t]
  \begin{center}
    \scalebox{0.27}{
        \includegraphics{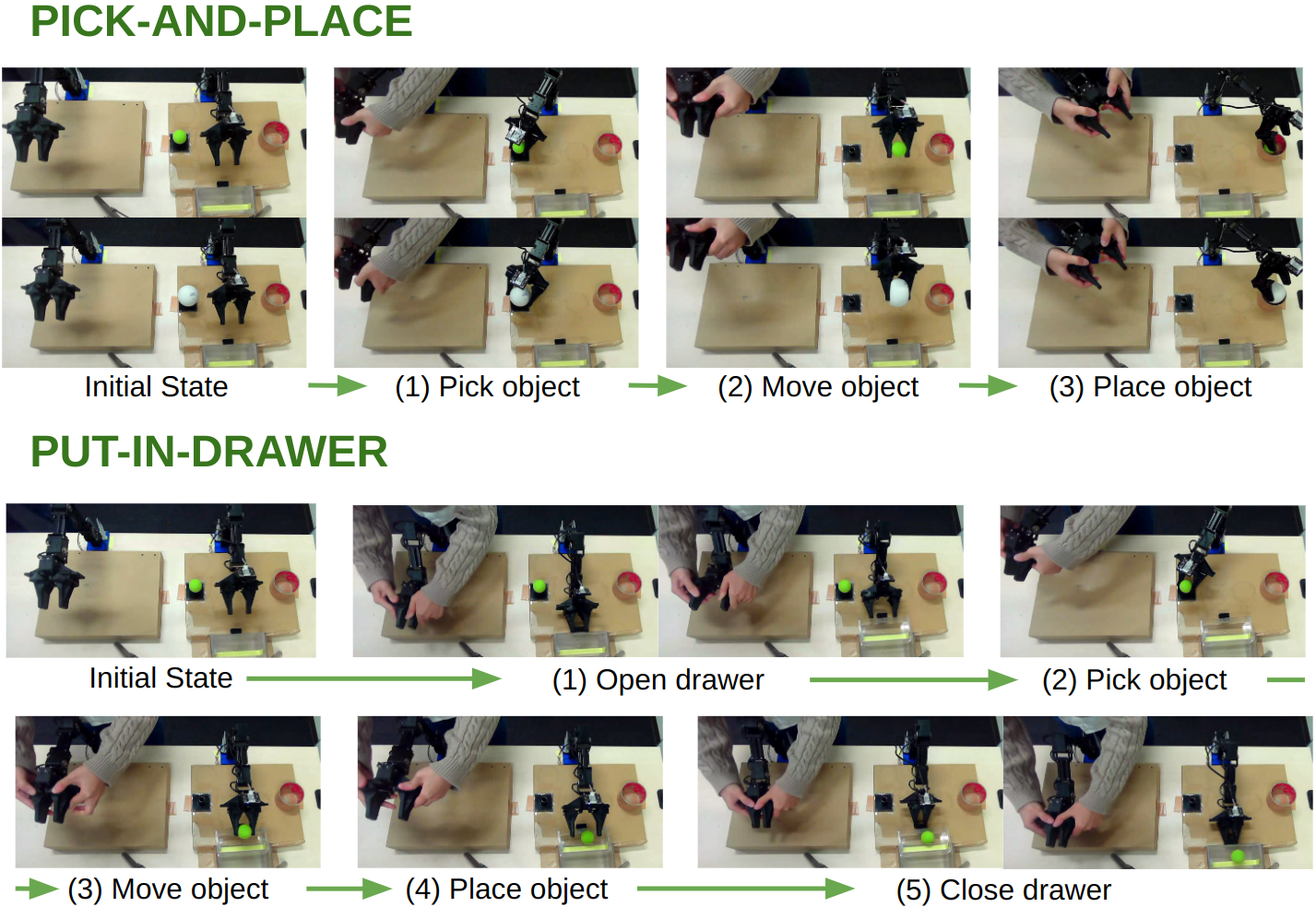}}    
  \caption{Snapshot of Data Collection}
  \label{fig:data_cllection}
\end{center}
\end{figure*}
\subsection{Training Dataset}
We collected joint angles, angular velocities, and torques data for a Leader-Follower robot's demonstration using a bilateral control system. The robot was controlled at a frequency of 1000Hz. Additionally, both the onboard hand RGB camera and the top RGB camera on the environmental side of the robot were operating at approximately 200Hz.
To align both sets of data with the system's operating cycle, we adjusted the data to 100Hz for use as training data. This was done because the model's inference cycle is approximately 100Hz.

For Pick-and-Place, 50 episodes were recorded using two different objects, a foam ball and a softball, with 25 trials for each. Each task was executed within a time frame ranging from 8.4 to 9.3 seconds per task. This resulted in a total of over 44,184 time steps in the dataset.

For Put-in-Drawer, 50 episodes were recorded, each lasting between 19.5 and 22.4 seconds, amounting to a total of 97,972 time steps in the data. As the human operator are unable to perform the task consistently in the exact same position, the policy does not learn a specific approach to solve the task. Instead, it considers how to perform the next step based on the observational data from images and joint data.

\subsection{Experimental Results}
We evaluated the performance of proposed method by comparing it without force control. For the pick-and-place task, we conducted trials on two trained objects, a foam ball and a softball, as well as seven untrained objects: a table tennis ball, an eye-cream package, Canele, a soccer ball, a plastic bell pepper, a honey bottle and a glue jar. Given the distinct characteristics of each object, as detailed in Table \ref{tbl:data_obj}, we evaluated the model's generalization and adaptability to various scenarios. Table \ref{tab:Experimental_Results-Pick} shows the percentage accuracy of the model across ten trials for each object. The comprehensive results reveal that while the model demonstrates high accuracy without force control, the integration of force metrics notably enhances its effectiveness.

For the proposed method, The proposed method exhibited a 100\% success rate in picking up balls of diverse sizes and shapes, including both trained objects and two untrained objects, the table tennis ball and eye cream. It also achieves high accuracy with the other five untrained objects— the Canele, soccer ball, honey bottle, plastic bell pepper, and glue jar, with success rates of 80\%, 90\%, 90\%, 80\%, and 80\%, respectively despite not having been trained on them before. This performance, achieved without prior training on these items, underscores the robot's capability to handle objects of varied shapes, colors, and states. 

In contrast, the model that does not incorporate force data seems to operate effectively with smaller objects, such as foam ball and table tennis ball, where the gripping location can be easily identified from the visuals provided by the gripper camera. However, its effectiveness significantly diminishes when dealing with larger or deformable objects, or those with irregular shapes, where the optimal gripping positions are less apparent. This variation in performance accentuates the vital role of force feedback in augmenting the model's adaptability to objects with intricate geometries and varying consistencies.

The most pronounced difference between the proposed model and the one without force data is observed in their handling of the eye cream and glue jar. The model excluding force data demonstrated a notably lower performance with these items, registering only a 50\% accuracy rate. Both the eye cream and glue jar share a common feature: they contain liquid, leading to an inconsistent weight distribution. This characteristic likely posed significant challenges for the model without force data, underscoring the necessity of incorporating force data for managing objects with dynamic weight characteristics.

\begin{figure}[t]
  \begin{center}
    \scalebox{0.36}{
        \includegraphics{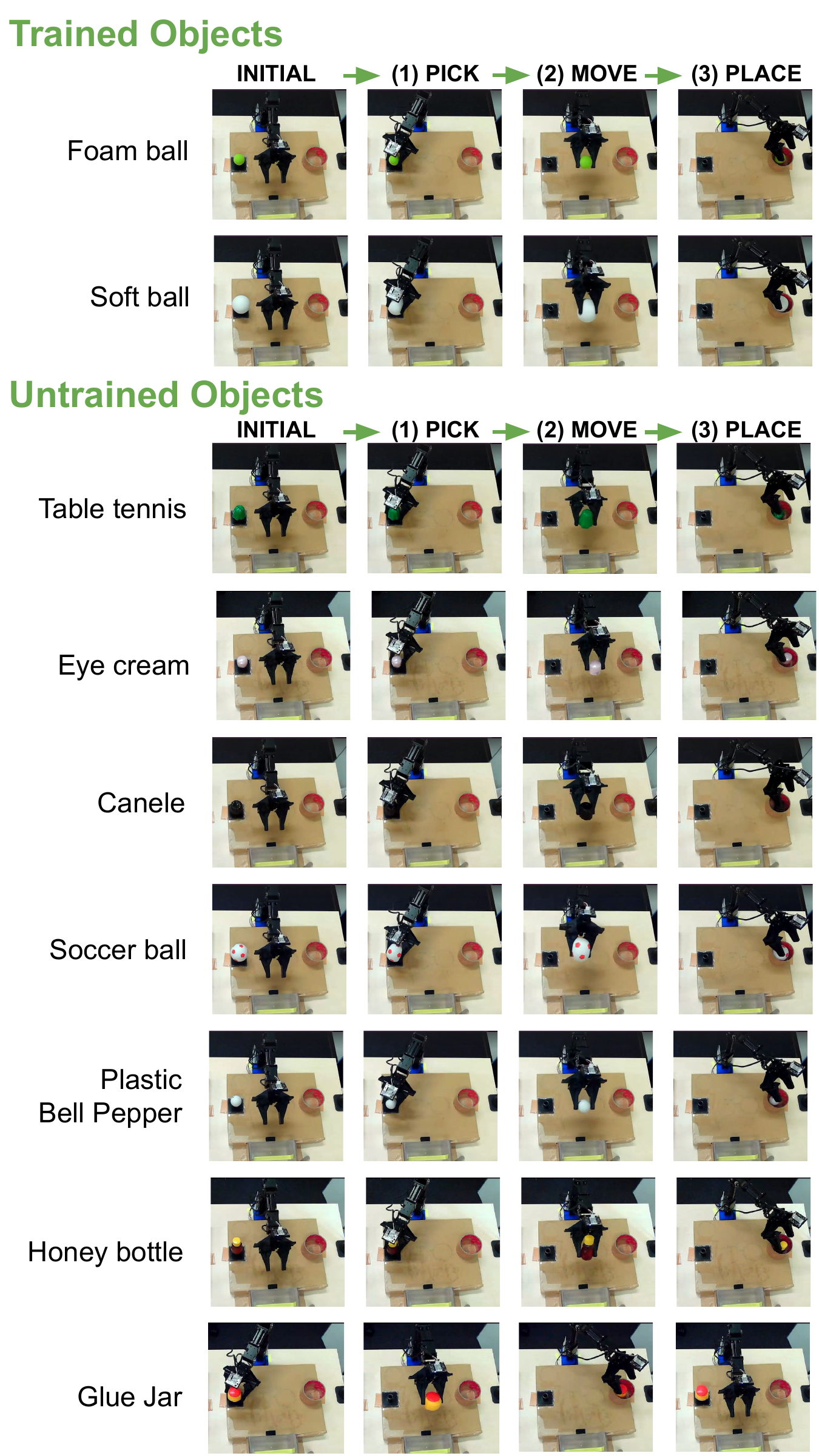}}    
  \caption{Experiment Result: Pick-and-Place (Proposed Method)}
  \label{fig:case1}
\end{center}
\end{figure}

\begin{table}[t]
    \centering
    \caption{Experimental Results: Pick-and-Place}
    \scalebox{0.88}{
    \begin{tabular}{c|cc|c|c|c|c}
        \hline
        \multicolumn{1}{c}{\multirow{2}{*}{Method}} & \multicolumn{2}{c}{\multirow{2}{*}{Objects}} &\multicolumn{4}{c}{Pick-and-Place}\\
        \multicolumn{1}{c}{}&\multicolumn{1}{c}{}&\multicolumn{1}{c}{}&\multicolumn{1}{c}{Pick}&\multicolumn{1}{c}{Move}&\multicolumn{1}{c}{Place}&Total\\\hline \hline
        
        \multirow{8}{*}{Bi-ACT} &\multirow{2}{*}{Trained}&Softball&100&100&100&100\\
        \multirow{8}{*}{(proposed)} &&Foam ball&100&100&100&100\\\cline{2-7}
        &\multirow{7}{*}{Untrained}&Table Tennis&100&100&100&100\\
        &&Eye Cream&100&100&100&100\\
        &&Canele&100&80&80&80\\
        &&Soccer Ball&90&90&90&90\\\
        &&Honey Bottle&90&90&90&90\\\
        &&Plastic Bell Pepper&80&80&80&80\\
        &&Glue Jar&80&80&80&80\\\hline
        \multirow{6}{*}{Bi-ACT} &\multirow{2}{*}{Trained}&Softball&80&80&80&80\\
        \multirow{6}{*}{(w/o Force)} &&Foam ball&100&100&100&100\\\cline{2-7}
        &\multirow{3}{*}{Untrained}&Table Tennis&100&100&100&100\\
        &&Eye Cream&70&50&50&50\\
        &&Canele&100&80&80&80\\
        &&Soccer Ball&90&90&80&80\\
        &&Honey Bottle&90&90&90&90\\\
        &&Plastic Bell Pepper&70&70&70&70\\
        &&Glue Jar&50&50&50&50\\\hline
    \end{tabular}
    }
    \label{tab:Experimental_Results-Pick}
\end{table}

For Put-in-Drawer, we performed the task on five trials as shown on the table \ref{tab:Experimental_Results-Drawer} in percentage. It could be seen that the robot arm were able to correctly perform the long-duration task without any significant errors, consistently achieving a 100\% success rate. This demonstrates the robustness and reliability of the model in executing complex tasks that require precise manipulation over extended periods. 
\begin{figure}[t]
  \begin{center}
    \scalebox{0.2}{
        \includegraphics{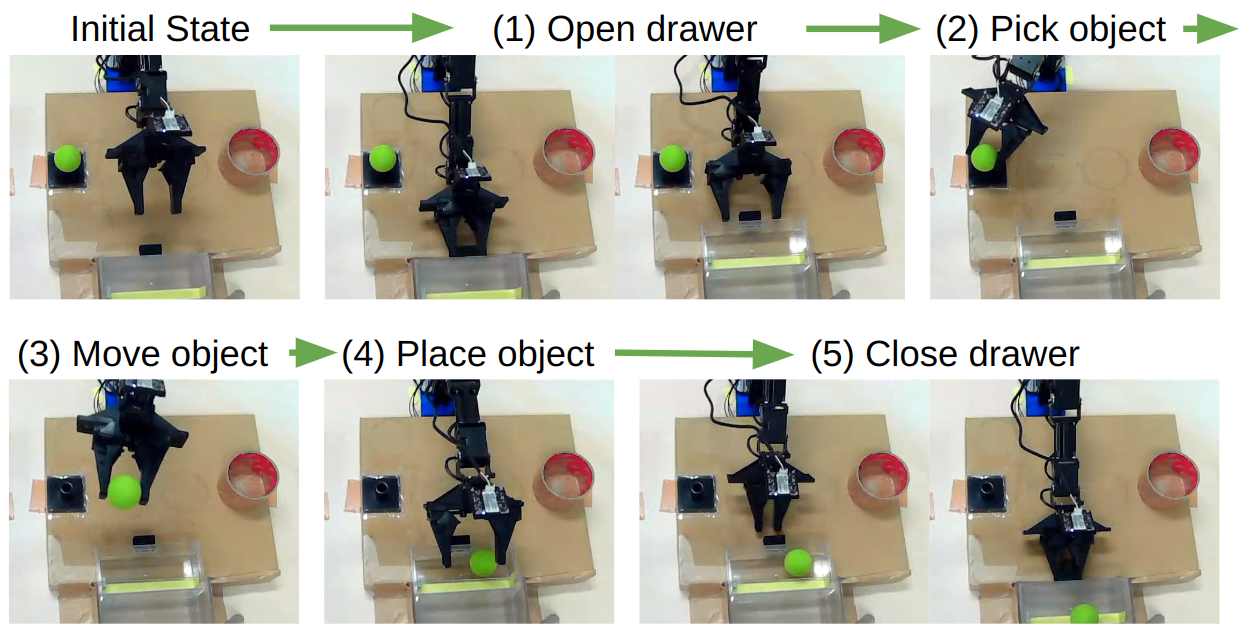}}    
  \caption{Experiment Result: Put-in-Drawer (Proposed Method)}
  \label{fig:case2}
\end{center}
\end{figure}

\begin{table}[t]
    \centering
    \caption{Experimental Results: Put-in-Drawer}
    \begin{tabular}{c|c|c|c|c|c}
        \hline 
        \multicolumn{1}{c}{\multirow{2}{*}{Method}} & \multicolumn{5}{c}{Put-in-Drawer}  \\
        \multicolumn{1}{c}{} &\multicolumn{1}{c}{Open}&\multicolumn{1}{c}{Pick}&\multicolumn{1}{c}{Move}&\multicolumn{1}{c}{Place}&\multicolumn{1}{c}{Close}\\\hline \hline
        Bi-ACT (proposed)&100&100&100&100&100\\\hline
    \end{tabular}
    \label{tab:Experimental_Results-Drawer}
\end{table}

\section{Conclusion}
This paper has proposed the Bilateral Control-Based Imitation Learning via Action Chunking with Transformer(Bi-ACT).
This approach intricately combines the robustness of bilateral control with the advanced computational prowess of the ACT architecture to process position and torque information for complex manipulative tasks.

We meticulously explored the application of the Bi-ACT model within the context of bilateral control.
Through this exploration, Bi-ACT has been shown to significantly outperform Bi-ACT without force methods in handling diverse datasets.
This advancement facilitates more effective learning and accurate prediction of robotic behavior in time series tasks, thereby improving the efficacy of autonomous manipulation.

The effectiveness of the proposed method was validated through real-world experiments.

In the future, we aim to refine and improve Bi-ACT as follows:
\begin{itemize}
\item {\it Robustness and Adaptability:} To ensure Bi-ACT's performance in diverse environments, we will work on making it more robust and adaptable. This includes addressing challenges related to varying lighting conditions, object recognition, and adapting to dynamic surroundings.

\item {\it Multimodal Sensory Integration:} We plan to expand Bi-ACT's capabilities by integrating multiple sensory inputs, such as vision, touch, and proprioception. This will enable more holistic perception and decision-making for the robot.

\item {\it Generalization Across Diverse Robotic Platforms:} The current version of Bi-ACT has undergone testing with a particular robotic arm setup.
We intend to broaden our assessments to encompass a more diverse array of robotic platforms, encompassing varying arm models and sensor configurations, in order to verify the versatility of the Bi-ACT framework.

\end{itemize}

\vfill


\begin{thebibliography}{1}
\bibliographystyle{IEEEtran}

\bibitem{IMIB2022yang}
S. Yang, W. Zhang, R. Song, W. Lu, H. Wang and Y. Li, "Explicit-to-Implicit Robot Imitation Learning by Exploring Visual Content Change," in IEEE/ASME Transactions on Mechatronics, vol. 27, no. 6, pp. 4920-4931, 2022.

\bibitem{IMI2023xu}
X. Xu, M. You, H. Zhou, Z. Qian and B. He, "Robot Imitation Learning From Image-Only Observation Without Real-World Interaction," in IEEE/ASME Transactions on Mechatronics, vol. 28, no. 3, pp. 1234-1244, 2023.

\bibitem{IMI2022ding}
J. Ding, T. L. Lam, L. Ge, J. Pang and Y. Huang, "Safe and Adaptive 3-D Locomotion via Constrained Task-Space Imitation Learning," in IEEE/ASME Transactions on Mechatronics, vol. 28, no. 6, pp. 3029-3040, 2023.

\bibitem{IMI2018yang}
C. Yang, C. Zeng, C. Fang, W. He and Z. Li, "A DMPs-Based Framework for Robot Learning and Generalization of Humanlike Variable Impedance Skills," in IEEE/ASME Transactions on Mechatronics, vol. 23, no. 3, pp. 1193-1203, 2018.

\bibitem{IMI2023zhou}
P. Zhou, X. Zhao, B. Tao and H. Ding, "Combination of Dynamical Movement Primitives With Trajectory Segmentation and Node Mapping for Robot Machining Motion Learning," in IEEE/ASME Transactions on Mechatronics, vol. 28, no. 1, pp. 175-185, 2023.

\bibitem{IMI2023franzese}
G. Franzese, L. d. S. Rosa, T. Verburg, L. Peternel and J. Kober, "Interactive Imitation Learning of Bimanual Movement Primitives," in IEEE/ASME Transactions on Mechatronics, 2023.

\bibitem{lfd2009argall}
B. D. Argall, S. Chernova, M. Veloso, and B. Browning, ``A survey of robot learning from demonstration,'' {\it Robotics and Autonomous Systems}, vol. 57, pp.~469-483, 2009.

\bibitem{lfd2022mukherjee}
D. Mukherjee, K. Gupta, L. H. Chang, and H. Najjaran, ``A Survey of Robot Learning Strategies for Human-Robot Collaboration in Industrial Settings,''  {\it Robotics and Computer-Integrated Manufacturing}, vol. 73, pp. 1-22, 2022.


\bibitem{tre2018Zhang}
T. Zhang, Z. McCarthy, O. Jow, D. Lee, X. Chen, K. Goldberg, and P. Abbeel, ``Deep Imitation Learning for Complex Manipulation Tasks from Virtual Reality Teleoperation,'' {\it Proceedings of IEEE International Conference on Robotics and Automation}, pp. 5628-5635, 2018.

\bibitem{tre2021tung}
A. Tung, J. Wong, A. Mandlekar, R. Mart´ın-Mart´ın, Y. Zhu, L. Fei-Fei, and S. Savarese, “Learning multi-arm manipulation through collaborative teleoperation,” {\it Proceedings of IEEE International Conference on Robotics and Automation}, pp. 9212-9219, 2021.


\bibitem{tre2018fan}
L. Fan, Y. Zhu, J. Zhu, Z. Liu, O. Zeng, A. Gupta, J. Creus-Costa, S. Savarese, and L. Fei-Fei, “SURREAL: Open-source reinforcement learning framework and robot manipulation benchmark,” {\it Proceedings of Conference on Robot Learning}, pp.~767-782, 2018.

\bibitem{act2023zhao}
T. Z. Zhao, V. Kumar, S. Levine and C. Finn, `` Learning Fine-Grained Bimanual Manipulation with Low-Cost Hardware,'' {\it arXiv}, 2304.13705, 2023.

\bibitem{mact2024fu}
Z. Fu, T. Z. Zhao, and C. Finn, ``Mobile ALOHA: Learning Bimanual Mobile Manipulation with Low-Cost Whole-Body Teleoperation,'' {\it arXiv}, 2401.02117, 2024.

\bibitem{ge2023swu}
P. Wu, Y. Shentu, Z. Yi, X. Lin, and P. Abbeel, "GELLO: A General, Low-Cost, and Intuitive Teleoperation Framework for Robot Manipulators" {\it arXiv}, 2309.13037, 2023.

\bibitem{IMIB2022sakaino}
S. Sakaino, K. Fujimoto, Y. Saigusa and T. Tsuji, ``Imitation Learning for Variable Speed Contact Motion for Operation up to Control Bandwidth,'' {\it IEEE Open Journal of the Industrial Electronics Society}, Vol. 3, pp. 116-127, 2022.

\bibitem{lstm}
S. Hochreiter and J. Schmidhuber, ``Long short-term memory'', {\it Neural Comput.}, vol. 9, no. 8, pp. 1735-1780, 1997.

\bibitem{con2021ke}
L. Ke, J. Wang, T. Bhattacharjee, B. Boots and S. Srinivasa, "Grasping with Chopsticks: Combating Covariate Shift in Model-free Imitation Learning for Fine Manipulation," {\it Proceedings of IEEE International Conference on Robotics and Automation}, pp.~6185-6191, 2021.

\bibitem{con2022swamy}
G. Swamy, S. Choudhury, J. A. Bagnell, and Z. S. Wu, "Causal Imitation Learning under Temporally Correlated Noise" {\it Proceedings of the International Conference on Machine Learning}, pp.~1~14, 2022.


\bibitem{IMIB2018adachi}
T. Adachi, K. Fujimoto, S. Sakaino and T. Tsuji, ``Imitation Learning for Object Manipulation Based on Position/Force Information Using Bilateral Control,'' {\it Proceedings of IEEE/RSJ International Conference on Intelligent Robots and Systems}, pp. 3648-3653, 2018.

\bibitem{IMIB2019fujimoto}
K. Fujimoto, S. Sakaino and T. Tsuji, ``Time Series Motion Generation Considering Long Short-Term Motion,'' {\it Proceedings of IEEE/RSJ International Conference on Intelligent Robots and Systems}, pp. 6842-6848,2019.

\bibitem{IMIB2020sasagawa}
A. Sasagawa, K. Fujimoto, S. Sakaino and T. Tsuji, ``Imitation Learning Based on Bilateral Control for Human–Robot Cooperation,'' {\it IEEE Robotics and Automation Letters}, vol. 5, no. 4, pp. 6169-6176, 2020.

\bibitem{IMIB2021sasagawa}
A. Sasagawa, S. Sakaino and T. Tsuji, ``Motion Generation Using Bilateral Control-Based Imitation Learning With Autoregressive Learning,'' {\it IEEE Access}, vol. 9, pp. 20508-20520, 2021.

\bibitem{IMIB2022sugiura}
Y. Saigusa, S. Sakaino and T. Tsuji, ``Imitation Learning for Nonprehensile Manipulation Through Self-Supervised Learning Considering Motion Speed,'' {\it IEEE Access}, vol. 10, pp. 68291-68306, 2022.

\bibitem{IMIB2022hayashi}
K. Hayashi, S. Sakaino and T. Tsuji, ``An Independently Learnable Hierarchical Model for Bilateral Control-Based Imitation Learning Applications,'' {\it IEEE Access}, vol. 10, pp. 32766-32781, 2022.

\bibitem{IMIB2023yamane}
K. Yamane, S. Sakaino and T. Tsuji, ``Soft and Rigid Object Grasping With Cross-Structure Hand Using Bilateral Control-Based Imitation Learning,'' {\it arXiv preprint}, arXiv:2311.09555, 2023.

\bibitem{IMIB2023kobayashi}
M. Kobayashi, T. Buamanee, Y. Uranishi, and H. Takemura, `` Research on Method for Generating Robot Arm Movements in Imitation Learning Based on Bilateral Control,'' {\it Proceedings of the Papers of Technical Meeting on Industrial Instrumentation and Control IEE Japan}, 	IIC-23033, pp.~35-38, 2023 (in Japanese).

\bibitem{TRANS2017vaswani}
A. Vaswani, N. Shazeer, N. Parmar, J. Uszkoreit, L. Jones, A. N. Gomez, L. Kaiser, and I. Polosukhin, ``Attention is All you Need,'' {\it Advances in Neural Information Processing Systems}, Vol.~30, 2017.



\bibitem{oneact2023george}
A. George and A. B. Farimani, `` One ACT Play: Single Demonstration Behavior Cloning with Action Chunking Transformers,'' {\it arXiv}, 2309.10175, 2023.



\bibitem{DOB}
K.Ohnishi, M. Shibata, T. Murakami, ``Motion Control for Advanced Mechatronics'', {\it IEEE/ASME Transactions on Mechatronics}, Vol.1,  No.1, pp.56–67, 1996.

\bibitem{RFOB}
T. Murakami, F. Yu, K. Ohnishi, ``Torque Sensorless Control in Multidegree-of-freedom Manipulator'', {\it IEEE Transactions on Industrial Electronics}, Vol.40, No.2, pp.259–265, 1993.
\end{thebibliography}
\end{document}